\documentclass[11pt]{article}

\usepackage[review]{acl}

\usepackage{times}
\usepackage{latexsym}

\usepackage[T1]{fontenc}
\usepackage[utf8]{inputenc}
\usepackage{microtype}
\usepackage{subfigure}
\usepackage{graphicx}
\usepackage{amssymb}
\usepackage{multirow}
\usepackage{pifont}
\usepackage{array}
\usepackage{booktabs}
\usepackage{amsmath}
\usepackage{amstext}
\usepackage{diagbox}
\usepackage{tabularx}
\usepackage{multirow}
\usepackage{enumitem}
\usepackage{CJKutf8}
\usepackage{bm}
\usepackage[ruled,vlined]{algorithm2e}
\setenumerate[1]{itemsep=0pt,partopsep=0pt,parsep=\parskip,topsep=0pt}
\setitemize[1]{itemsep=0pt,partopsep=0pt,parsep=\parskip,topsep=0pt}
\setdescription{itemsep=0pt,partopsep=0pt,parsep=\parskip,topsep=0pt}
\title{MulZDG: Multilingual Code-Switching Framework for Zero-shot Dialogue Generation}

\author{Yongkang Liu, Shi Feng, Daling Wang\thanks{Corresponding author} , Yifei Zhang \\
  Northeastern University, Shenyang, China \\
  \texttt{misonsky@163.com} \\
  \texttt{\{fengshi, wangdaling, zhangyifei\}@cse.neu.edu.cn} \\
  }

\begin{document}
\maketitle

\begin{abstract}
Building dialogue generation systems in a zero-shot scenario remains a huge challenge, since the typical zero-shot approaches in dialogue generation rely heavily on large-scale pre-trained language generation models such as GPT-3 and T5. The research on zero-shot dialogue generation without cumbersome language models is limited due to lacking corresponding parallel dialogue corpora. In this paper, we propose a simple but effective \textbf{Mul}tilingual learning framework for \textbf{Z}ero-shot \textbf{D}ialogue \textbf{G}eneration (dubbed as MulZDG) that can effectively transfer knowledge from an English corpus with large-scale training samples to a non-English corpus with zero samples. Besides, MulZDG can be viewed as a multilingual data augmentation method to improve the performance of the resource-rich language. First, we construct multilingual code-switching dialogue datasets via translation utterances randomly selected from monolingual English datasets. Then we employ MulZDG to train a unified multilingual dialogue model based on the code-switching datasets. The MulZDG can conduct implicit semantic alignment between different languages. Experiments on DailyDialog and DSTC7 datasets demonstrate that MulZDG not only achieve competitive performance under zero-shot case compared to training with sufficient examples but also greatly improve the performance of the source language.
\end{abstract}
\section{Introduction}
The success of neural models and the emergence of large-scale dialogue datasets have greatly advanced the research of dialog generation~\cite{serban2016building,serban2017hierarchical,huang2020challenges,meng2020dukenet}. The open-domain dialogue systems aim to generate more informative and fluent responses~\cite{ke2018generating,zhang2020dialogpt,bao2020plato,meng2021initiative}, which are widely used in various applications such as emotional companionship, mental health support, and social chatbots.
\begin{figure}[h]
\centering
\includegraphics[width=0.85\linewidth]{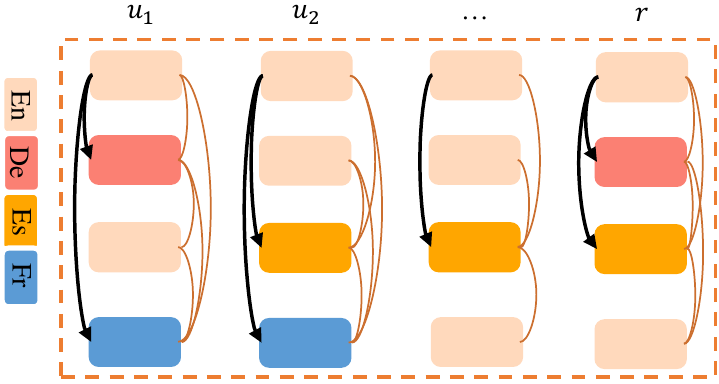}
\caption{Multilingual semantic alignment map in zero-shot case. One color represents one language. English is the source language and other languages are target languages. $u_i$ represents the \textit{i}-th utterance in the conversation history and $r$ represents the responses. Black arrows represent translating utterances from English into other languages. Solid brown lines represent implicit semantic alignments between different languages.}
\label{multi_semantic}
\end{figure}

Although achieving promising performance, most existing dialogue generation systems~\cite{zhang2020dialogpt,bao2020plato,li2020optimus,floridi2020gpt} rely on a considerable amount of data resource, such as DialoGPT~\cite{zhang2020dialogpt}. In practice, the dialogue corpus for many languages is unavailable, which limits the usefulness of dialogue systems for low-resource or even zero-resource languages. Hence it is important to design approaches that can effectively transfer knowledge from the source language with sufficient resources to a target language with zero training samples.

The pre-trained language models have been proved to be very effective in dialogue generations including zero-shot scenarios. Most existing zero-shot dialogue generation approaches usually directly employ large-scale pre-trained generative language models such as GPT-3~\cite{brown2020language}, T5~\cite{raffel2020exploring}, or conduct secondary pre-training on target language based on language models~\cite{ebrahimi2021adapt,kim2021model}. Although these methods can handle the issue of zero-shot generation, the cost of pre-training is unaffordable.

In the similar task of NMT (neural machine translation), zero-shot generation methods are mainly introducing one additional language between source and target language~\cite{johnson2017google,zheng2017maximum,artetxe2018unsupervised,cheng2019joint,liu2020multilingual}, which can achieve indirect semantic mapping between the source and target language through the additional intermediate language. The nature of translation is the semantic mapping between source and target languages while there is no similar semantic mapping between dialogue history and response. Therefore, directly transferring the zero-shot approaches in NMT to the dialogue generation task is infeasible.

The zero-shot multilingual understanding tasks usually employ code-switching approach to achieve semantic alignment between different languages~\cite{liu2020attention,chapuis2021code,qin2021cosda}. The way of code-switching can conduct implicit semantic alignment without relying on parallel corpus pairs. Inspired by these studies, we employ the code-switching method to transfer the knowledge of dialogue history in English to other target languages which have no training examples. We follow previous work~\cite{chapuis2021code} on multilingual representation to employ code-switching at the utterance level, although code-switching at the word or span level is more common~\cite{banerjee2018dataset,bawa2020multilingual,dougruoz2021survey}.

Based on the code-switching method, we propose a simple but effective \textbf{Mul}tilingual learning framework for \textbf{Z}ero-shot \textbf{D}ialogue \textbf{G}eneration (dubbed as MulZDG) that can effectively transfer knowledge from English corpus with large-scale training samples to non-English corpus with zero samples. Specifically, we first construct code-switch languages using the NMT system and bilingual dictionary~\cite{pan2021contrastive}. As shown by the black arrows in Figure~\ref{multi_semantic}, we randomly select utterances from dialogue history to translate into other target languages. For each target language, we construct code-switching corpus containing source and target languages, respectively.
Then we employ MulZDG based on an encoder-decoder structure to train a unified multilingual dialogue generation system, which can be applied in the source language and other target languages with no training samples. MulZDG with a multi-task structure can generate responses with different languages according to specific input prompts. MulZDG can conduct implicit semantic alignments through task sharing mechanism between different languages, as shown by solid brown lines in Figure~\ref{multi_semantic}. To summarize, we make the following contributions:
\begin{itemize}[leftmargin=*]
\item We propose a simple but effective multilingual framework, MulZDG, which can effectively transfer the knowledge from the source language with large-scale training samples to target languages with zero samples.
\item We present a data augmentation method for multilingual code-switching, which can enhance the the performance of source language.
\item We construct multilingual code-switching dialogue datasets from English dialogue datasets DailyDialog and DSTC7, and release the multilingual versions datasets\footnote{https://github.com/misonsky/MultilingualDatasets}.
\end{itemize}
\section{RELATED WORK}
\subsection{Dialogue Generation}
Dialogue generation systems aim to produce informative and fluent responses and have attracted great attention in academia. Early 
studies usually adopt the methods of NMT based on an Encoder-Decoder network to generate responses~\cite{sordoni2015hierarchical,serban2016building}. However, these methods often generate dull and generic responses. To tackle this problem, memory mechanism~\cite{wu2019global,zhang2020memory,tian2020response} and attention mechanism~\cite{zhang2019recosa} are introduced into dialogue modeling successively. Although many approaches have been proposed, there are still remarkable gaps between responses generated by neural models and those from humans~\cite{holtzman2019curious}. Large-scale pre-trained generative models~\cite{zhang2020modeling,ling2021context,wang2020topic} have greatly facilitated the development of dialogue generation task. Although achieving promising performance, these methods rely on a large number of training examples, which greatly limits the usability of dialogue systems. In this paper, we propose a multilingual framework that can work in zero-shot case.
\subsection{Zero-shot Learning}
Zero-shot Learning for dialogue generation tasks refers to building a dialogue generation system without available training samples~\cite{floridi2020gpt}. Most existing zero-shot methods of dialogue generation rely on large-scale pre-trained generative models~\cite{lewis2020bart,zhang2020dialogpt,floridi2020gpt}, such as GPT-3~\cite{floridi2020gpt}. These methods require huge computing resources, which hinders the usableness of dialogue systems. On similar machine translation tasks, zero-shot methods mainly include triangular and multilingual NMT systems. The triangular NMT systems~\cite{zheng2017maximum,cheng2019joint} build a triangular translation case by adding an intermediate language and multilingual systems~\cite{johnson2017google,liu2020multilingual} mix 
multiple different language pairs to achieve semantic alignment between different languages under zero-shot case by building implicit 
triangular systems. The nature of these methods is the semantic mapping between different languages, which is consistent with the translation task. However, there is no similar semantic mapping between dialogue history and response. In this paper, we propose a simple but effective multilingual dialogue generation framework based on code-switching languages.
\begin{figure*}[h]
\centering
\includegraphics[width=0.90\textwidth]{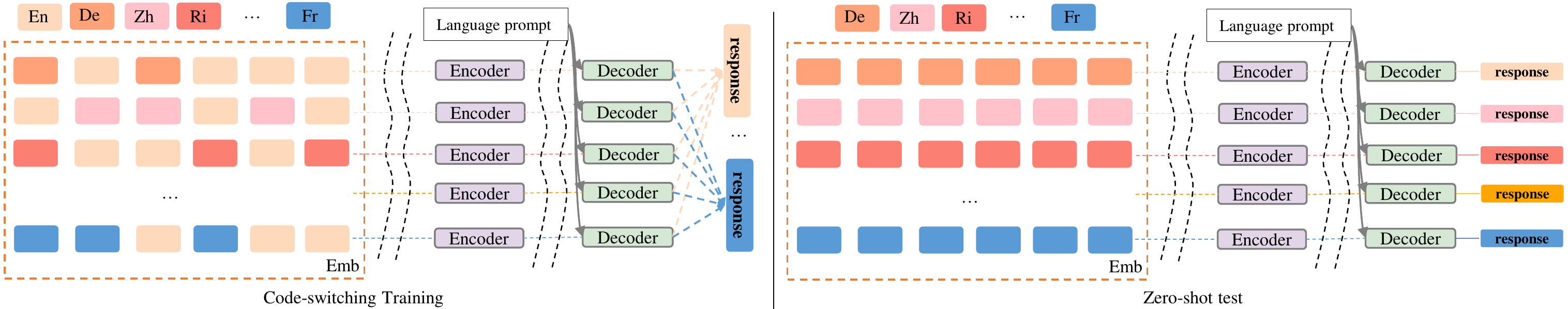}
\caption{The framework of MulZDG.}
\label{framework}
\end{figure*}
\section{METHODOLOGY}
\subsection{Problem Formalization}
Given a source language ($S$), the goal is to build a unified dialogue generation system based on the source language that can be applied to target languages ($T$). In this paper, the source language \textit{S} is English. The $T$ include: Zh (Chinese), De (German), Ru (Russian), Es (Spanish), Fr (French) and It (Italian). An instance in dialogue dataset can be represented as $(C,R)$ where $C = \{u_1,u_2,...,u_n\}$ with $n$ utterance represents the context of dialogue. $u_i$ represents the \textit{i}-th utterance. And $R$ represents the corresponding response. Based on the \textit{S}, we automatically build code-switching languages that include \textit{T} through random translations. The construction process of code-switching languages are shown in Algorithm~\ref{alg:cs}, where $D_{T} = \{D_{en_{zh}}, D_{en_{de}}, D_{en_{ru}}, D_{en_{es}}, D_{en_{fr}}, D_{en_{it}}\}$. The $D_{en_{tar}}$ represents code-switching languages composed of English and $tar$.
\begin{algorithm}[t]
  \caption{Code-Switching Languages.}
  \label{alg:cs}
  \KwIn{Corpus of \textit{S}: $D_{S}$; The target language set $T$; Bilingual Dictionary: $Dic$;}
  \KwOut{Code-switching languages:$D_{T}$.}
  \BlankLine
  Initialize $D_{T} = \emptyset$; 

  \ForEach{$(C,R)$ in $D_{s}$}{
    random sample utterances: $U \sim P(C,R)$;
    
    \ForEach{target language $t_i$ in $T$}{
      Replace tokens for each utterance in $U$ by dictionary $Dic(S \rightarrow t_i)$: $\tilde{U}$;

      Translate $\tilde{U}$ by NMT system: $\hat{U}$;

      Add updated $(C,R))$ by $\hat{U}$ to $D_{T}$;
    }
  }
\end{algorithm}
\subsection{Multi-task MulZDG Framework}
As shown in Figure~\ref{framework}, MulZDG mainly contains three layers: (i) Embedding Layer; (ii) Encoder Layer; (iii) Decoder Layer. The embedding layer is responsible for mapping each word in $D_{T}$ to a vector space using a pre-trained Glove~\cite{pennington2014glove} embedding model. The encoder layer is responsible for capturing the semantic representation of the dialogue context. And the decoder layer is responsible for generating the probability distribution of response for the dialogue context. The decoder in MulZDG can generate responses in different languages through different prompt tokens. The code-switching languages in $D_{T}$ share the same encoder and decoder, which can conduct semantic alignment between different languages through task sharing mechanism. In the training phase, we employ code-switching languages to train MulZDG. In the testing phase, MulZDG takes monolingual dialogue samples as input and generates responses in the corresponding target language. 

MulZDG is a multi-task framework, which supports networks with RNN and transformer as the backbone. The inputs of frameworks are utterance-level code-switching languages. Different inputs in $D_{T}$ share the same encoder and decoder. The encoder can be hierarchical or non-hierarchical structures. The framework with a non-hierarchical encoder is just like the seq2seq arthitecture~\cite{sutskever2014sequence} which consists of an encoder RNN and a decoder RNN. The framework with a hierarchical encoder is just like HRED~\cite{serban2016building} which consists of a hierarchical encoder and a decoder.

Encoders can be hierarchical or non-hierarchical. The framework with a non-hierarchical encoder is just like the seq2seq arthitecture~\cite{sutskever2014sequence} which consists of an encoder and a decoder. The framework with a hierarchical encoder is just like HRED~\cite{serban2016building} which consists of a hierarchical encoder and a decoder. The structure of encoder can be RNN or transformer structure. Next, we briefly introduce both hierarchical and non-hierarchical frameworks.
\subsection{Non-hierarchical MulZDG}
The non-hierarchical MulZDG is a simple seq2seq structure, which has a shared encoder and a decoder module. The encoder is responsible for encoding the dialogue context and the decoder is responsible for generating the response for the dialogue context. Encoder-Decoder structure based on RNN~\cite{sutskever2014sequence} and Transformer~\cite{vaswani2017attention} are the most classic representative models.

\textbf{RNN Encoder-Deocder}
In this paper, we employ GRU~\cite{cho2014learning} as the concrete implementation for RNN.
We concatenate code-switching language utterances in dialogue context into a consecutive tokens sequence, and add a special start 
symbol [SOS] and end symbol [EOS] to each tokens sequence. The last hidden state of encoder is denoted as $h_{\ell}$, which is considered as the dialogue context summary. $\ell$ represents the length of sequence. Finally, we employ another GRU to decode the probability distribution of response.
\begin{equation}
  P_{\theta}(y | h_{\ell}) = \prod \limits_{j=2}^m P_{\theta}(y_j | y_1,...,y_{j-1},h_{\ell})P(y_1)
\end{equation}
where $\theta$ represents the parameters of GRU and $y$ represents the response sequence to be decoded.

\textbf{Transformer Encoder-Deocder} Each layer of transformer~\cite{vaswani2017attention} encoder is composed of a multi-head self-attention mechanism and a position-wise fully connected feed-forward network. A residual connection~\cite{he2016deep} is employed around each layer, followed by layer normalization~\cite{ba2016layer}. In addition to the two sub-layers in each encoder layer, the decoder layer possesses another multi-head attention calculating the weight distribution over the output of the encoder based on predicted sequences. Note that in addition to the word embedding information, position encoding information is required in the embedding layer.
\subsection{Hierarchical MulZDG}
The encoder of hierarchical MulZDG possesses two basic components: utterance encoder and context encoder, which is like the HRED~\cite{serban2016building} architecture. The utterance encoder is responsible for capturing the features of each utterance and the context encoder is in charge of distilling the dependencies between different utterances. The framework MulZDG supports the implementation of RNNs and transformers structure. In this paper, we employ GRU as the implementation of RNNs.

\textbf{Hierarchical Transformer Structure} 
Note that positional encoding is required for the implementation of the transformer. The utterance encoder and decoder layer share the same positional encoding. The outputs of utterance encoder based on transformer can be denoted as $u_i=\{e_{sos},e_1,e_2,...,e_{m},e_{eos}\}$ for i-th utterance, where $m$ represents the length of utterance. We employ an average pooling operation to get a fixed-dimensional representation for every utterance. We add positional encoding to $h_i$ and the context encoder based on transformer is responsible for capturing the global dialogue context information $\hat{H} = \{\hat{h_1},\hat{h_2},\cdots,\hat{h_n}\}$. Then the decoder layer decodes the probability distribution of response over $\hat{H}$.
\begin{equation}
  h_i = \frac{1}{m}\sum_{j=0}^{m} e_j
\end{equation}
\section{Experiments} 
\subsection{Datasets}
 In this paper, we select English datasets DailyDialog~\cite{li2017dailydialog} and DSTC7~\cite{galley2018end} as the source languages and generate code-switching training examples based on source training datasets through Algorithm~\ref{alg:cs}. We directly translate the test datasets into the corresponding target languages. Note that we employ different translation systems (described in Appendix~\ref{sec:NMT}).

 \textbf{DailyDialog} is a multi-turn dialogue dataset about our daily life, which consists of 11,118 context-response pairs for training, 1,000 pairs for validation, and 1,000 pairs for testing. The proportion of non-English utterances in code-switching dataset is De-30.186\% in $D_{en_{de}}$, Es-30.134\% in $D_{en_{es}}$, Fr-30.096\% in $D_{en_{fr}}$, It-30.293\% in $D_{en_{it}}$, Ru-30.199\% in $D_{en_{ru}}$, Zh-30.109\% in $D_{en_{zh}}$. In the experiment we abbreviate it as Daily.

 \textbf{DSTC7} is a multi-turn dialogue dataset from social media data, which consists of 76,590 context-response pairs for training, 17,870 pairs for validation, and 1,710 pairs for testing. The proportion of non-English utterances in code-switching dataset is De-27.661\% in $D_{en_{de}}$, Es-27.545\% in $D_{en_{es}}$, Fr-27.639\% in $D_{en_{fr}}$, It-27.611\% in $D_{en_{it}}$, Ru-27.786\% in $D_{en_{ru}}$, Zh-27.565\% in $D_{en_{zh}}$.
\begin{table}[h]
\tiny
\centering
\begin{tabular}{p{.55cm}p{.4cm}p{.12cm}p{.16cm}p{0.7cm}p{0.16cm}p{0.7cm}p{1.4cm}}
\toprule
Models & Datasets & Types & PPL & BL-1/2 & RL & Dist-1/2 & Embed A/E/G \\ 
\midrule
\multirow{4}{*}{HRED} 
  &\multirow{2}{*}{Daily}
     & En &127.7& 29.87/24.05& 35.54& 12.56/44.55& 80.55/82.17/64.22 \\
     & & Aug & 122.4 & 34.47/28.31 &39.11 &13.33/45.67 &82.66/82.88/65.76  \\
  \cline{2-8}
  &\multirow{2}{*}{DSTC7}
    & En &116.9& 26.73/17.38& 29.03& 5.34/24.52& 78.98/84.78/61.66\\
    & & Aug &116.4&27.92/18.77& 29.47&7.55/25.16& 80.02/84.86/59.66 \\
\midrule
\multirow{4}{*}{VHRED} 
  &\multirow{2}{*}{Daily}
     & En&123.3 & 34.69/25.77& 40.77& 13.77/45.56& 84.74/86.17/69.77 \\
     & & Aug&123.1& 35.93/27.35& 41.88 &14.57/46.16& 86.88/87.94/71.23 \\
  \cline{2-8}
  &\multirow{2}{*}{DSTC7}
     & En & 127.7& 25.54/14.44& 25.47& 7.49/32.93& 77.77/85.68/57.98 \\
    & & Aug & 123.5& 27.05/16.92& 26.52& 8.03/33.74& 79.39/86.24/58.17 \\
\midrule
\multirow{4}{*}{Trans}
  &\multirow{2}{*}{Daily}
     & En &143.3&22.86/14.77 &28.55 &10.36/33.63& 79.96/80.06/63.15 \\
    & & Aug &141.2&23.89/16.17& 30.35& 11.79/35.64& 81.23/81.22/65.45 \\
  \cline{2-8}
  &\multirow{2}{*}{DSTC7}
      & En &163.4& 22.77/19.74 &21.57 &6.77/34.56& 78.32/82.56/56.89 \\
      & & Aug &158.9& 24.47/23.44& 23.67& 7.98/35.02& 80.19/84.47/56.88 \\
\midrule
\multirow{4}{*}{HTrans} 
  &\multirow{2}{*}{Daily}
    & En &146.7& 23.76/15.61& 27.3&  9.96/35.79& 80.26/79.44/62.62 \\
    & & Aug &133.5& 24.57/17.03& 28.94& 11.12/36.98& 82.33/82.16/63.76 \\
  \cline{2-8}
  &\multirow{2}{*}{DSTC7}
    & En &162.5& 23.54/18.68& 23.35& 7.32/35.66& 80.02/82.14/64.33 \\
    & & Aug &153.3& 25.78/20.37& 25.76& 8.69/37.84& 82.43/84.34/65.54 \\
 \bottomrule
\end{tabular}
\caption{Performance of models based on data augmentation using multilingual code-switching of monolingual source. Eng represents that models is only trained on monolingual source language. Aug represents that models is trained on mixed languages including source and multilingual code-switching based on MulZDG. BL stands for BLEU and RL represents Rouge-L. All values are multiplied by 100.}
\label{tab:En}
\end{table}
\begin{table}[h]
\tiny
\centering
\begin{tabular}{p{.55cm}p{.4cm}p{.12cm}p{.16cm}p{0.7cm}p{0.16cm}p{0.7cm}p{1.4cm}}
\toprule
Models & Datasets & Types & PPL & BL-1/2 & RL & Dist-1/2 & Embed A/E/G \\ 
\midrule
\multirow{4}{*}{HRED} 
  &\multirow{2}{*}{Daily}
    & De &125.4 &25.89/20.17&35.72 &15.63/49.55&78.23/86.86/62.42 \\
    & & Multi &125.9& 23.77/20.33& 34.31& 15.34/48.89&78.03/86.54/59.97 \\
  \cline{2-8}
  &\multirow{2}{*}{DSTC7}
      & De &124.6&21.68/11.89&21.71&7.35/35.23&77.19/87.63/58.41 \\
      & & Multi &119.7&23.85/12.87& 23.77& 7.66/33.67&77.37/87.57/60.02\\
\midrule
\multirow{4}{*}{VHRED} 
  &\multirow{2}{*}{Daily}
  & De &121.2 & 27.14/20.44 &  34.77 &  16.53/51.94&   77.56/87.26/62.09\\ 
  & & Multi &124.6 & 25.87/20.67 &  33.56&   15.63/51.32&   77.04/87.56/61.42\\
  \cline{2-8}
  &\multirow{2}{*}{DSTC7}
  & De &125.3 &  22.74/12.07&   23.69 &  7.68/36.29&   78.34/87.44/58.62\\
  & & Multi &123.4&   20.04/11.34&   23.56 &  6.89/36.52&   76.88/87.12/57.96\\ 
\midrule
\multirow{4}{*}{Trans} 
  &\multirow{2}{*}{Daily}
     & De &150.2&   20.06/12.38&   24.96&   9.35/34.56&   74.63/78.88/60.76\\
    & & Multi &146.7&   18.67/11.89&   22.65&   8.98/33.63&   74.53/77.98/59.67 \\
  \cline{2-8}
  &\multirow{2}{*}{DSTC7}
    & De &156.2&   20.55/11.37&   22.76&   5.64/32.19&   76.54/85.55/55.43\\
    & & Multi &156.8&   18.77/10.23&   20.87&   4.88/30.87&  75.87/84.71/55.13\\
\midrule
\multirow{4}{*}{HTrans} 
  &\multirow{2}{*}{Daily}
  & De &156.7&   21.33/11.66&   25.22&   8.96/33.19&  75.61/77.98/61.86\\
  & & Multi &157.8&   19.65/10.98&   24.33&   8.87/33.65&  74.78/79.13/59.43\\ 
  \cline{2-8}
  &\multirow{2}{*}{DSTC7}
    & De &157.7&   19.88/11.24&   23.45&   5.77/33.47&   75.96/86.73/56.47\\
    & & Multi &161.5&   18.78/11.42&   21.77&   5.33/32.19&   75.47/86.33/56.44\\
 \bottomrule
\end{tabular}
\caption{Zero-shot results of models on German. De represents that models is only trained on German training set. Multi represents that models is only trained on code-switching languages based on MulZDG.} 
\label{tab:De}
\end{table}
\begin{table}[h]
\tiny
\centering
\begin{tabular}{p{.55cm}p{.4cm}p{.12cm}p{.16cm}p{0.7cm}p{0.16cm}p{0.7cm}p{1.4cm}}
\toprule
Models & Datasets & Types & PPL & BL-1/2 & RL & Dist-1/2 & Embed A/E/G \\ 
\midrule
\multirow{4}{*}{HRED} 
  &\multirow{2}{*}{Daily}
     & Es &124.1&25.49/20.75 &36.97& 17.68/52.65& 77.69/84.21/63.44\\
    & & Multi &126.7& 26.45/20.55 &35.98& 17.33/50.56& 76.53/83.68/63.56\\
  \cline{2-8}
  &\multirow{2}{*}{DSTC7}
    & Es &119.6&  25.39/14.47& 26.94& 7.58/30.13& 77.99/87.32/61.24\\
    & & Multi &118.4& 25.47/15.76& 29.23& 7.23/30.67& 77.68/86.77/62.13\\
\midrule
\multirow{4}{*}{VHRED} 
  &\multirow{2}{*}{Daily}
  & Es &122.1& 29.14/23.04& 35.99& 18.82/55.37& 77.79/85.52/62.28\\
  & & Multi&126.7& 30.88/22.52& 34.76& 17.35/54.88& 77.23/84.56/62.03\\
  \cline{2-8}
  &\multirow{2}{*}{DSTC7}
  & Es &114.5& 26.24/14.42& 27.82& 8.78/32.45& 78.68/87.88/59.77\\
  & & Multi &123.3& 26.45/17.39& 31.19& 7.49/32.19& 78.97/87.57/63.64\\
\midrule
\multirow{4}{*}{Trans} 
  &\multirow{2}{*}{Daily}
  & Es &157.7& 19.06/3.65& 22.06& 8.35/30.65& 73.61/77.67/58.36\\
  & & Multi &153.7& 18.76/12.98& 22.05& 8.14/31.48& 73.32/76.88/59.54\\
  \cline{2-8}
  &\multirow{2}{*}{DSTC7}
  & Es &123.3& 24.38/15.52& 26.44& 7.53/30.62& 77.66/86.48/60.12\\
  & & Multi &125.7& 24.33/15.43& 24.55& 7.43/30.88& 77.42/86.04/62.33\\
\midrule
\multirow{4}{*}{HTrans} 
  &\multirow{2}{*}{Daily}
  & Es &154.6& 18.93/10.57& 21.76& 8.07/29.33& 72.55/75.69/56.38\\
  & & Multi &156.3& 17.97/9.47&  21.62& 8.33/27.74& 72.76/75.4/56.12\\
  \cline{2-8}
  &\multirow{2}{*}{DSTC7}
  & Es &127.8&  23.87/14.33& 26.56& 7.44/29.48& 77.87/85.46/59.56\\
  & & Multi & 131.2& 23.19/12.59& 24.89& 7.95/29.65& 76.46/83.12/57.15\\
 \bottomrule
\end{tabular}
\caption{Zero-shot results of models on Spanish. Es represents models is trained on monolingual Spanish.}
\label{tab:Es}
\end{table}
\subsection{Baselines}
MulZDG is a general multilingual learning framework that can be applied to various dialogue generation models. We select several representative models in this paper. 

\textbf{HRED}~\cite{serban2016building} is a hierarchical encoder-decoder structure with a hierarchical encoder (including utterance encoder and context encoder) and a decoder. Hred employs shared encoders to encode each utterance separately. 

\textbf{VHRED}~\cite{serban2017hierarchical} is a hierarchical encoder-decoder structure with a hierarchical encoder (incuding utterance encoder and context encoder) and a decoder based on variational mechanism. VHRED can generate long outputs with better use of contextual information via latent variables. 

\textbf{Transformer}~\cite{vaswani2017attention} is a encoder-decoder structure with multi-head attention mechanism. The inputs of the Transformer is a consecutive word sequence concatenated all utterances. In all experimental tables, we abbreviate it as \textbf{Trans}.

\textbf{HTransformer}~\cite{santra2020hierarchical} is a hierarchical encoder-decoder structure with a hierarchical encoder (including utterance encoder and context encoder) and a decoder. Htransformer encodes each utterance separately. In all experimental tables, we abbreviate it as \textbf{HTrans}.
\subsection{Implementation Details}
We implement our MulZDG using Tensorflow 2. The word embedding size and hidden size are all set to 512. We employ Adam optimizer~\cite{kingma2014adam} to train all models. For the models HRED and VHRED we set the learning rate to 0.001. For the Transform and HTransformer, we set the learning rate to 0.0001. For the models HRED and VHRED, we set the number of encoder and decoder layers to 1. For the Transformer and HTransformer, the number of encoder and decoder layers is 3. And the number HRED for Transformer and HTransformer is 8. We employ the word segmenttation of BERT~\cite{devlin2019bert} and the vocabulary of multilingual BERT as the unified vocabulary. The batch size is 128 for HRED and VHRED, and we set it to 512 for Transformer and HTransformer. The maximum epochs are set to 200. We employ GloVe to train a unified multilingual embedding vectors representation for embedding-based metrics based on multilingual corpora. We do not remove unknown tokens when computing embedding-based metrics, and the vectors of all unknown tokens are initialized to zero vector. 
\begin{table}[h]
\tiny
\centering
\begin{tabular}{p{.55cm}p{.4cm}p{.12cm}p{.16cm}p{0.7cm}p{0.16cm}p{0.7cm}p{1.4cm}}
\toprule
Models & Datasets & Types & PPL & BL-1/2 & RL & Dist-1/2 & Embed A/E/G \\ 
\midrule
\multirow{4}{*}{HRED}
  &\multirow{2}{*}{Daily}
  & Fr &122.3& 25.36/19.89& 35.15& 13.59/43.68& 80.44/87.12/64.43\\
  & & Multi &124.6& 25.12/18.78& 35.04& 13.35/42.45& 79.31/87.09/62.21\\
  \cline{2-8}
  &\multirow{2}{*}{DSTC7}
    & Fr &111.3& 26.42/16.82& 27.18& 4.7/21.18& 80.84/88.36/62.59 \\
    & & Multi &116.8& 26.54/15.96& 25.47& 5.43/24.33& 80.45/88.31/60.61\\
\midrule
\multirow{4}{*}{VHRED} 
  &\multirow{2}{*}{Daily}
  & Fr &132.2& 25.64/19.49& 34.97& 15.42/45.22& 79.86/87.77/63.83\\
  & & Multi &135.4& 24.96/18.87& 32.43& 13.88/43.88& 79.89/87.57/65.13\\
  \cline{2-8}
  &\multirow{2}{*}{DSTC7}
  & Fr &107.6&  28.88/19.36& 29.24& 5.85/26.22& 81.18/88.35/63.8\\
  & & Multi &112.4& 26.93/18.47& 27.56& 5.77/26.6&  80.39/88.26/61.24\\
\midrule
\multirow{4}{*}{Trans} 
  &\multirow{2}{*}{Daily}
  & Fr &163.7& 16.66/12.06& 25.3&  7.47/29.34& 75.66/77.14/59.78\\
  & & Multi &166.5& 15.76/11.56& 24.54& 7.23/29.87& 76.32/77.65/57.62\\
  \cline{2-8}
  &\multirow{2}{*}{DSTC7}
  & Fr &122.3& 25.64/16.47& 28.04& 5.12/25.63& 79.94/87.57/61.25\\
  & & Multi &124.2& 23.65/14.67& 27.33& 5.34/24.77& 79.54/86.32/60.43\\
\midrule
\multirow{4}{*}{HTrans} 
  &\multirow{2}{*}{Daily}
  & Fr &164.4& 16.61/9.49&  22.82& 7.57/31.98& 75.12/78.06/60.08\\
  & & Multi &165.4& 15.44/8.54&  21.65& 6.77/31.33& 75.54/78.67/60.36\\
  \cline{2-8}
  &\multirow{2}{*}{DSTC7}
  & Fr &125.6& 24.87/17.77& 26.89& 4.77/22.34& 79.56/88.45/60.56\\
  & & Multi &127.3& 24.18/16.28& 24.86& 4.53/22.54& 79.89/87.23/62.65\\
 \bottomrule
\end{tabular}
\caption{Zero-shot results of models on French. Fr represents models is trained on monolingual French.}
\label{tab:Fr}
\end{table}
\begin{figure}[h]
\centering
\includegraphics[width=0.80\linewidth]{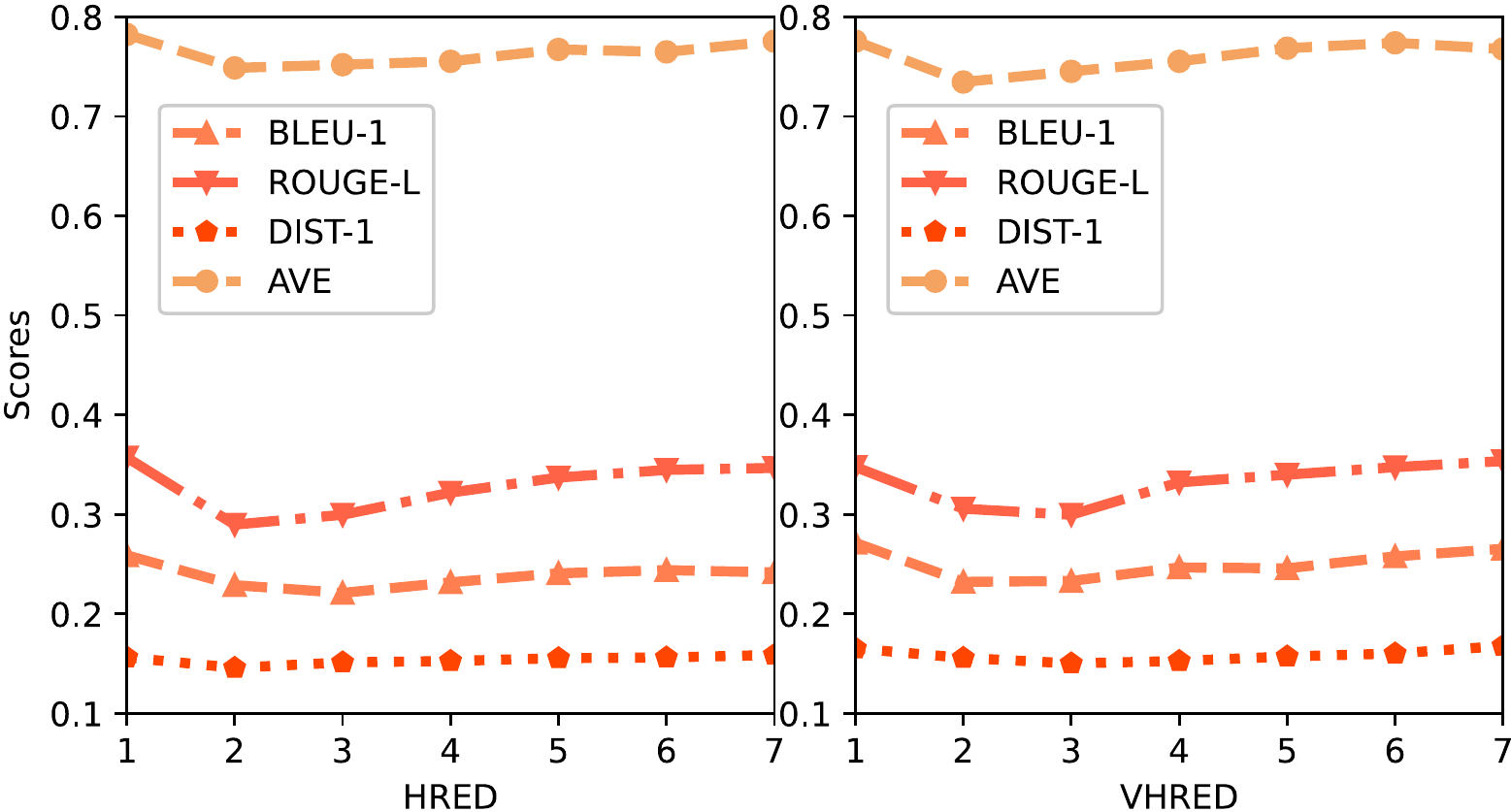}
\caption{The impact of the number of multilingual code-switching languages on models performance for German. The '1' in abscissa represents that models are trained on monolingual German. Number of languages is greater than '1' represents models is trained on multilingual code-switching languages based on MulZDG.}
\label{number_languages}
\end{figure}
\begin{table}[h]
\tiny
\centering
\begin{tabular}{p{.55cm}p{.4cm}p{.12cm}p{.16cm}p{0.7cm}p{0.16cm}p{0.7cm}p{1.4cm}}
\toprule
Models & Datasets & Types & PPL & BL-1/2 & RL & Dist-1/2 & Embed A/E/G \\ 
\midrule
\multirow{4}{*}{HRED} 
  &\multirow{2}{*}{Daily}
  & It &126.7& 25.51/20.01& 35.84& 18.07/53.16& 76.34/84.54/61.98\\
  & & Multi &126.7& 25.01/19.23& 33.98& 18.33/53.25& 74.78/84.13/60.3\\
  \cline{2-8}
  &\multirow{2}{*}{DSTC7}
  & It &108.5& 24.26/16.11& 27.26& 7.56/34.53& 75.69/87.07/58.73\\
  & & Multi &111.2& 22.65/14.52& 25.54& 7.13/32.15& 73.85/87.04/56.96\\
\midrule
\multirow{4}{*}{VHRED} 
  &\multirow{2}{*}{Daily}
  & It & 123.7& 26.99/20.82& 34.69& 19.64/57.47& 76.56/85.38/60.77\\
  & & Multi &124.4& 25.11/19.14& 32.68& 17.72/55.89& 74.39/84.33/59.31\\
  \cline{2-8}
  &\multirow{2}{*}{DSTC7}
  & It &114.6& 28.66/19.04& 29.42& 8.57/35.12& 80.99/88.31/63.96\\
  & & Multi &125.4& 26.87/18.44& 28.49& 7.54/35.12& 78.54/87.19/63.56\\
\midrule
\multirow{4}{*}{Trans} 
  &\multirow{2}{*}{Daily}
  & It &156.5& 18.11/15.72& 27.15& 8.75/32.34& 75.37/79.67/62.08\\
  & & Multi & 160.5& 16.65/14.36& 27.42& 7.87/31.33& 74.36/78.23/61.32\\
  \cline{2-8}
  &\multirow{2}{*}{DSTC7}
  & It &123.3& 25.55/17.57& 26.44& 6.54/32.12& 75.09/86.88/57.97\\
  & & Multi &124.3& 23.55/16.55& 24.87& 6.12/30.35& 75.01/85.32/57.89\\
\midrule
\multirow{4}{*}{HTrans}
  &\multirow{2}{*}{Daily}
  & It &153.5& 19.43/17.56& 29.3&  9.05/30.34& 76.22/80.89/63.28\\
  & & Multi &154.7& 17.87/18.04& 28.88& 9.87/30.54& 76.86/80.23/62.67\\
  \cline{2-8}
  &\multirow{2}{*}{DSTC7}
  & It &124.5& 26.76/18.97& 25.67& 7.56/34.56& 76.11/85.12/57.44\\
  & & Multi &123.6& 24.88/18.95& 24.77& 7.45/32.17& 76.59/85.07/56.21\\
 \bottomrule
\end{tabular}
\caption{Zero-shot results of models on Italian. It represents models is trained on monolingual Italian.}
\label{tab:It}
\end{table}
\subsection{Evaluation metrics} 
To compare different models, we employ both automatic metrics and human evaluations. \textbf{Automatic Metrics}: We employ perplexity (PPL) and distinct 1/2 (Dist.1/2) following previous sdudies~\cite{zhang2018personalizing,zheng2020pre,song2021bob}. Lower perplexity means more reliable model. Distinct 1/2~\cite{li2015diversity} are the ratio of distinct uni-grams / bi-grams. Higher distinct means better diversity of responses generated by the model. We also employ BLEU~\cite{papineni2002bleu} and ROUGE-L~\cite{lin2004rouge} (abbreviated as RL) for evaluating response generation. BLEU and ROUGE-L metrics evaluate the response based on o-occurrence properties of tokens. Embedding-based metrics (Average, Exterma and Greedy)~\cite{liu2016not,xu2018better,sedoc2019chateval} can reflect the quality of the generated responses at the semantic level. \textbf{Human Evaluation}: We further conduct human evaluations to assess the proposed learning framework. We select Chinese and English dialogue systems for human evaluation on DailyDialog and DSTC7. We ask three crowd-sourced graduate students to evaluate the quality of generated responses for 100 randomly sampled input contexts. We request annotators to choose a preferred response, or vote a tie, considering the following aspects of response quality: fluency, informativeness, coherence, and engagingness.
\begin{table}[h]
\tiny
\centering
\begin{tabular}{p{.55cm}p{.4cm}p{.12cm}p{.16cm}p{0.7cm}p{0.16cm}p{0.7cm}p{1.4cm}}
\toprule
Models & Datasets & Types & PPL & BL-1/2 & RL & Dist-1/2 & Embed A/E/G \\ 
\midrule
\multirow{4}{*}{HRED} 
  &\multirow{2}{*}{Daily}\
    & Ru & 123.8& 28.43/22.16& 37.51& 21.36/53.56& 78.07/85.83/64.34\\
    & & Multi &127.6& 27.57/21.57& 36.65& 20.54/52.15& 76.32/84.84/64.37\\
  \cline{2-8}
  &\multirow{2}{*}{DSTC7}
  & Ru &119.5& 28.21/14.25& 29.95& 10.22/35.21& 80.68/86.61/64.17\\
  & & Multi &116.5& 27.99/18.19& 28.93& 9.32/35.12& 81.38/88.68/63.51\\
\midrule
\multirow{4}{*}{VHRED} 
  &\multirow{2}{*}{Daily}
  & Ru &134.3& 27.96/21.88& 35.66& 21.46/55.88& 78.66/85.63/63.22\\
  & & Multi &133.2& 25.43/20.47& 33.23& 20.13/53.17& 76.98/85.65/63.22\\
  \cline{2-8}
  &\multirow{2}{*}{DSTC7}
  & Ru &103.6& 29.49/17.02& 34.25& 11.15/36.61& 79.96/86.13/66.41\\
  & & Multi &107.7& 27.77/16.87& 34.77& 10.86/34.65& 80.03/86.88/65.74\\
\midrule
\multirow{4}{*}{Trans} 
  &\multirow{2}{*}{Daily}
  & Ru &144.6& 21.53/13.77& 27.68 &9.57/34.93& 78.37/81.55/64.5\\
  & & Multi &147.9& 20.54/13.28& 27.43 &8.76/33.98& 78.54/80.13/62.19\\
  \cline{2-8}
  &\multirow{2}{*}{DSTC7}
  & Ru &123.3& 25.66/13.27& 27.84& 9.46/32.67& 78.88/84.55/65.14\\
  & & Multi &125.4& 24.65/12.09& 25.48& 8.91/30.54& 77.99/84.33/64.87\\
\midrule
\multirow{4}{*}{HTrans} 
  &\multirow{2}{*}{Daily}
  & Ru &143.1& 20.77/14.65& 28.49& 9.68/35.20&  78.75/81.33/63.05\\
  & & Multi &147.6& 18.44/13.76& 26.35& 9.02/33.89& 78.96/79.57/63.06\\
  \cline{2-8}
  &\multirow{2}{*}{DSTC7}
  & Ru & 127.8& 24.87/14.77& 26.46& 9.13/31.66& 77.88/83.67/65.03\\
  & & Multi &128.6& 22.76/13.56& 24.32& 8.25/29.78& 77.54/81.98/65.23\\
 \bottomrule
\end{tabular}
\caption{Zero-shot results of models on Russian. Ru represents models is trained on monolingual Russian.}
\label{tab:Ru}
\end{table}
\subsection{Effectiveness of Data Augmentation}
In particular, models based on transformer do not employ pre-trained language model as the initial checkpoint and train from scratch. In addition, the performance of models based on transformer perform worse than RNN in experiments due to the limited training datasets.

Table~\ref{tab:En} reports the results of models on DailyDialog and DSTC7 using monolingual English corpus and data augmentation based on multilingual code-switching languages. We can observe that the performances of models are greatly improved when using data augmentation based on multilingual code-switching. Specifically, the performances of models using data augmentation with multilingual code-switching is 0.2\% to 13.2\% on PPL, 0.81\% to 4.38\% higher on BLEU-1, 1.4\% to 3.7\% higher on BLEU-2, 0.59\% to 2.41\% higher on Rouge-L, 0.54\% to 1.43\% higher on dist-1, 0.46\% to 2.64\% higher on dist-2, and 1.27\% to 2.41\% higher on average embedding compared with models using monolingual English corpus. The data augmentation approaches with multilingual code-switching can enhance the representation ability of models and improve the quality of the responses through learning common features between different languages.
\begin{figure}[h]
\centering
\includegraphics[width=0.80\linewidth]{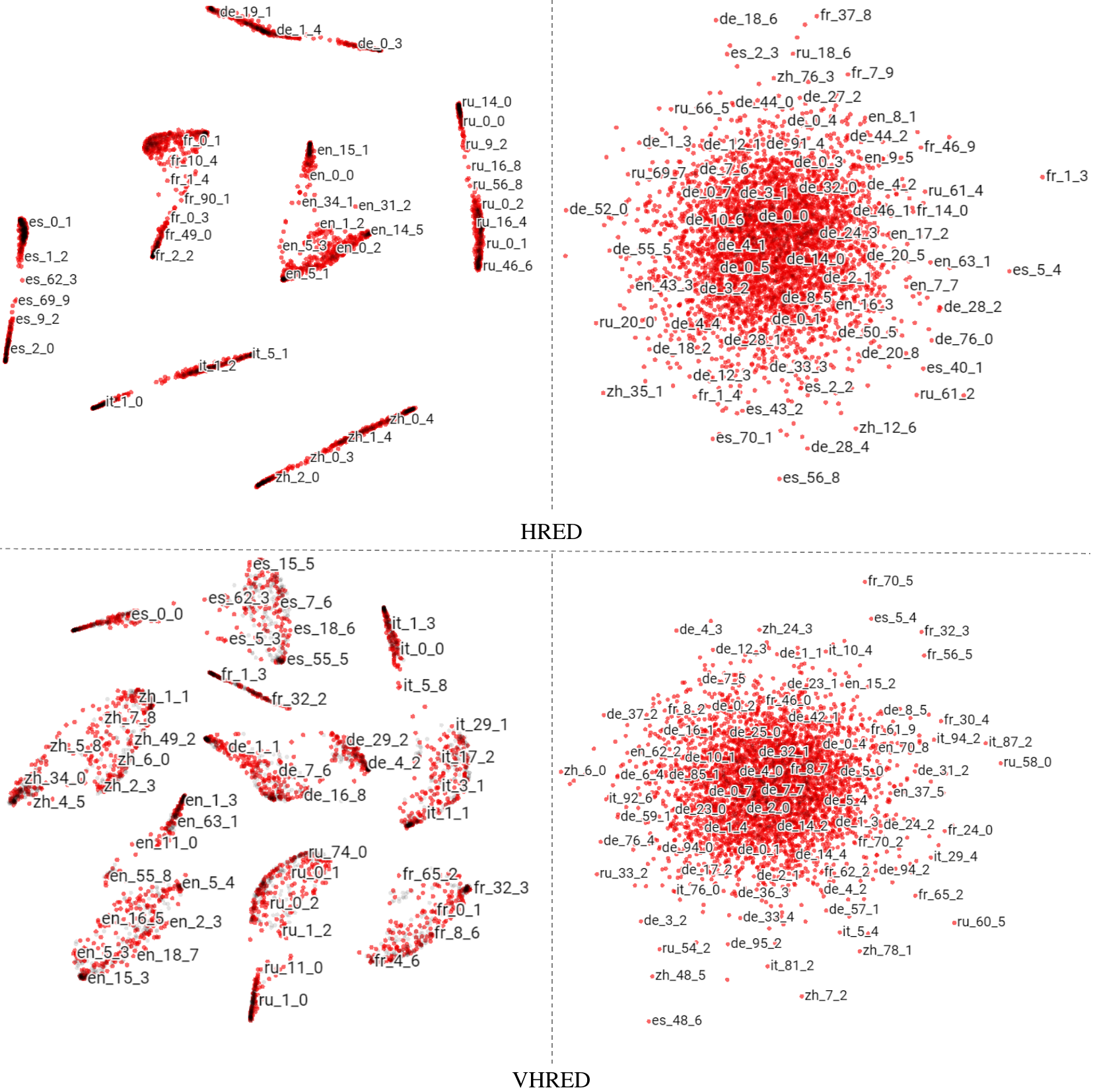}
\caption{Semantic alignment visualization. The left presents the visualization of utterance vectors for models trained on monolingual language. The right demonstrates the visualization of utterance vectors for models trained on multilingual code-switching languages based on MulZDG. The label $lan$-$N_1$-$N_2$ represents the number of the utterance, where $lan$ stands for language, $N_1$ stands for the number of data sample and $N_2$ indicates the utterance number of utterances in dialogue history.}
\label{mul_semantic_tsn}
\end{figure}
\begin{table}[h]
\tiny
\centering
\begin{tabular}{p{.55cm}p{.4cm}p{.12cm}p{.16cm}p{0.7cm}p{0.16cm}p{0.7cm}p{1.4cm}}
\toprule
Models & Datasets & Types & PPL & BL-1/2 & RL & Dist-1/2 & Embed A/E/G \\ 
\midrule
\multirow{4}{*}{HRED} 
  &\multirow{2}{*}{Daily}
  & Zh &127.4& 23.52/19.52& 30.83& 9.5/43.37& 83.05/83.77/68.42\\
  & & Multi &124.3& 21.34/18.11& 28.33& 9.05/42.07& 83.56/82.75/66.88\\
  \cline{2-8}
  &\multirow{2}{*}{DSTC7}
  & Zh &108.1& 18.07/12.68& 26.78& 4.06/25.24& 85.32/86.27/73.63\\
  & & Multi &116.7& 17.91/11.55& 24.83& 4.04/23.81& 85.8/86.34/72.29\\
\midrule
\multirow{4}{*}{VHRED} 
  &\multirow{2}{*}{Daily}
  & Zh &125.6& 24.54/20.03& 31.28& 11.22/46.08& 83.17/84.43/67.77\\
  & & Multi &127.3& 24.57/19.09& 29.88& 9.97/44.13& 82.18/84.32/66.49\\
  \cline{2-8}
  &\multirow{2}{*}{DSTC7}
  & Zh &110.4& 19.92/13.32& 27.61& 4.56/26.57& 85.79/86.54/72.01\\
  & & Multi &116.5& 17.98/12.65& 26.17& 4.36/24.79& 84.15/85.32/72.4\\
\midrule
\multirow{4}{*}{Trans} 
  &\multirow{2}{*}{Daily}
  & Zh &165.7& 15.55/11.54& 24.38& 7.89/33.47& 79.32/80.46/66.21\\
  & & Multi &167.4& 15.13/10.33& 22.99& 6.54/31.44& 76.98/78.54/66.21\\
  \cline{2-8}
  &\multirow{2}{*}{DSTC7}
  & Zh &123.4& 18.77/12.45& 26.58 &4.15/24.46&83.56/85.67/71.77\\
  & & Multi &123.3& 18.04/11.87& 25.37& 4.54/23.87& 84.13/83.51/70.7\\
\midrule
\multirow{4}{*}{HTrans}
  &\multirow{2}{*}{Daily}
  & Zh &166.8& 16.92/11.83& 25.66& 8.61/34.21& 79.92/81.51/66.54\\
  & & Multi &167.7& 15.66/11.09& 24.76& 7.96/33.2& 78.99/80.23/66.46\\
  \cline{2-8}
  &\multirow{2}{*}{DSTC7}
  & Zh &134.5&  19.04/12.44& 25.67& 4.32/25.67& 84.39/85.19/70.65\\
  & & Multi &135.4& 18.45/11.32&  23.87&  4.32/25.33& 83.79/83.99/69.69\\
 \bottomrule
\end{tabular}
\caption{Zero-shot results of models on Chinese. Zh represents models is trained on monolingual Chinese.}
\label{tab:Zh}
\end{table}
\subsection{Zero-shot Dialogue Generation}
Different from Table~\ref{tab:En}, Table~\ref{tab:De} to Table~\ref{tab:Zh} report the results of zero-shot generation using MulZDG based on multilingual code-switching languages. The performances of models trainging on multilingual code-switching languages can achieve competitive results under zero-shot case compared with on corresponding monolingual language. On German DailyDialog and DSTC7, the performances of models under zero-shot case is 0.1\% higher on PPL, 1.23\% lower on BLEU-1, 0.93\% on Rouge-L, 0.71\% on dist-1 and 0.51\% on average embedding compared with models training on monolingual corpus according to Table~\ref{tab:De}. On French DailyDialog and DSTC7, the performances of models under zero-shot case is 2.9\% lower on PPL, 0.93\% lower on BLEU-1, 1.34\% on Rouge-L, 0.27\% on dist-1 and 0.16\% on average embedding compared with models training on monolingual corpus according to Table~\ref{tab:Fr}. On Chinese datasets, the performances of models under zero-shot case is 2.1\% lower on PPL, 0.91\% lower on BLEU-1, 1.57\% on Rouge-L, 0.44\% on dist-1 and 0.62\% on average embedding compared with models training on monolingual corpus according to Table~\ref{tab:Zh}.

We can observe the similar results on other languages. MulZDG adopts a multi-task approach to generate responses in different languages. On the one hand, sharing the structure between multiple tasks is benefical for models to exploit the common features between different languages. On the other hand, sharing task mechanism between multilingual code-switching languages is beneficial to enhance the semantic alignment ability of models between different languages.
\begin{table}[h]
\tiny
\centering
\begin{tabular}{lcccc|ccc} 
\toprule
\multirow{3}{*}{Models} & \multirow{3}{*}{Languages} & \multicolumn{6}{c}{Datasets} \\
\cline{3-8}
 & &\multicolumn{3}{c}{Daily (\%)} & \multicolumn{3}{c}{DSTC7 (\%)} \\ 
 \cline{3-8}
 & & Win & Tie & Loss & Win & Tie & Loss \\ \hline
 \multirow{3}{*}{HRED} 
  &En & 30 & 44 & 26 & 33 & 36 & 31 \\ \cline{2-8}
  &Fr & 23 & 56 & 21 & 23 & 54 & 23 \\
  &Zh & 31 & 35 & 34 & 32 & 33 & 35 \\ \midrule
 \multirow{3}{*}{VHRED}
  &En & 38 & 32 & 30 & 30 & 45 & 25 \\ \cline{2-8}
  &Fr & 16 & 63 & 21 & 11 & 82 & 7 \\
  &Zh & 36 & 29 & 35 & 30 & 37 & 33 \\
\bottomrule
\end{tabular}
\caption{Human evaluation on DailyDialog and DSTC7 (multilingual VS monolingual). On English corpus, we compare the the performances of models training using data augmentation of multilingual code-switching with monolingual English corpus. On other languages, we compare the performances of models under zero-shot using multilingual code-switching languages with corresponding monolingual corpus.}
\label{tab:human_evaluation}
\end{table}

\subsection{Impact of Multilingualization}
To explore the effect of multilingual code-switching languages, we conduct further experiments. Figure~\ref{number_languages} demonstrates the effect of the number of languages on models performance. We select German as the target language (other languages are available), HRED and VHRED as the tested models. The performances of bilingual code-switching languages on HRED and VHRED are dramatically lower than models trained on the corresponding monolingual training set. However, the performances of models are gradually improved with the number of languages increases. We can only add up to seven languages due to the limitations of our constructed corpus. We can conclude that MulZDG fails to work well in zero-shot case when the number of languages is small according to ~\ref{number_languages}. More constructive conclusions require further experimental evaluation in the case of more languages in the future, such as hundreds of languages.
\subsection{Multilingual Mechanism Analysis}
We conduct extensive experiments to explore how the multilingual mechanism works. We select 1,000 multilingual parallel examples from DailyDialog, about 4,536 utterances in total and visualize the representations of these examples in HRED and VHRED based on MulZDG trained on different monolingual languages and multilingual code-switchinig languages, respectively. Figure~\ref{mul_semantic_tsn} presents the results of the utterances representation visualization on HRED and VHRED. The representations of utterances based on different monolingual languages are clustered into different language categories while representations of utterances on multilingual code-switchinig languages are clustered together according to semantics. Utterances in different languages expressing same semantics will be clustered together, which demonstrates that MulZDG will do semantic alignment between different languages. This phenomenon presents that MulZDG pays attention to the common features between different languages.
\subsection{Human Evaluation}
Although automatic evaluation metrics have been shown to be reliable, we still conduct human evaluations to confirm the validity of the MulZDG. We compare the performance of models training on multilingual code-switching languages and on corresponding monolingual language.
Table~\ref{tab:human_evaluation} reports the results of human evaluation of HRED and VHRED on three languages (i.e., English, French and Chinese).
We can observe that the performances of models using data augmentation with multilingual code-switching is on average 6.0\% higher on DailyDialog and 3.5\% higher on DSTC7 than using monolingual English corpus. Besides, the performance of models under zero-shot is on average 5\% lower on DailyDialog and 2\% lower on DSTC7 than using coresponding monolingual corpus. These results demonstrate that multilingual code-switching framework can not only be considered as a data augmentation method but also be employed to zero-shot dialogue generation.
\section{Conclusion and Future Work}
In this paper, we propose a simple but effective multilingual framework: MulZDG, which can not only be used as an approach of data augmentation but also be used to zero-shot dialogue generaton. Besides, we release the multilingual versions of DailyDialog and DSTC7 datasets. In the future, we will explore the working mechanism and effect of large-scale multilingual code-switching languages (e.g., hundreds of languages) in the problem of zero-shot dialogue generation.
% , which contains seven languages: English, Chinese, German, Russian, Spanish, French and Italian
\bibliography{anthology,custom}
\bibliographystyle{acl_natbib}

\appendix
\section{Appendix}
\label{sec:appendix}
\subsection{NMT Systems} \label{sec:NMT}
We simulate distribution differences between training and test datasets through using different translation systems. The NMT systems employed for the training set are \textit{Helsinki-NLP/opus-mt-en-tar}\footnote{https://huggingface.co/Helsinki-NLP/}. 'tar' represents the target languages, which includes Chinese, German, Russian, Spanish, French and Italian. And we employ T5-base\footnote{https://huggingface.co/t5-base} to translate the test datasets into German and French. For Russian, Chinese, Spanish and Italian we employ WMT19-en-ru\footnote{https://huggingface.co/facebook/wmt19-en-ru}, WMT-en-zh\footnote{https://huggingface.co/liam168/trans-opus-mt-en-zh}, mbart-en-es\footnote{https://huggingface.co/mrm8488/mbart-large-finetuned-opus-en-es-translation} and Google NMT system, respectively.

\begin{table*}[h]
\centering
\begin{tabular}{ccc}
\toprule
Models & Types & Responses \\ 
\midrule
\multirow{2}{*}{HRED} 
  & En &I like travelling. like my experience and i have a good idea. \\
  & Aug &I like travelling. I travel frequently.  \\
\midrule
\multirow{2}{*}{VHRED} 
  & En &i like sports. I like travelling. \\
  & Aug & I like travelling. I'll do it next time. \\ 
\midrule
\multicolumn{2}{c}{Gold Response} &  Yes, I like travelling. I am young, and unmarried. It’s no problem for me to travel frequently. \\
 \bottomrule
\end{tabular}
\caption{Case study for data augmentation on Daily dataset. The English context is "\textit{What are your personal weaknesses? I’m afraid I’m a poor talker. I’m not comfortable talking with the people whom I have just met for the first time. That is not very good for business, so I have been studying public speaking. Are you more of a leader or a follower? I don’t try to lead people. I’d rather cooperate with everybody, and get the job done by working together. Do you think you can make yourself easily understood in English? Yes, in most circumstances. Are you available for travel?}"}
\label{tab:CaseEn}
\end{table*}

\end{document}